\documentclass{article}
\usepackage{spconf,amsmath,graphicx}
\usepackage{amsmath,amssymb,amsfonts}
\usepackage{bm}
\usepackage{booktabs}
\usepackage{enumitem}
\usepackage{algorithm}
\usepackage{algorithmic}

\def\method{DAGRL}
\title{Unsupervised Domain Adaptive Graph Classification}
\name
 {Siyang Luo$^{1\dag}$, Ziyi Jiang$^{2\dag}$, Zhenghan Chen$^{3}$\sthanks{Contact author: 1979282882@pku.edu.cn}, Xiaoxuan Liang$^4$}
\address{$^1$Shanghai Infortech Software Development Co.,Ltd, Shanghai, China\\
	$^2$New York University Shanghai, Shanghai, China\\
	$^3$Peking University, Beijng, China\quad $^4$University of Massachusetts Amherst, Amherst MA, USA   }
\begin{document}
%
\maketitle
\begin{abstract}

Despite the remarkable accomplishments of graph neural networks (GNNs), they typically rely on task-specific labels, posing potential challenges in terms of their acquisition. Existing work have been made to address this issue through the lens of unsupervised domain adaptation, wherein labeled source graphs are utilized to enhance the learning process for target data. However, the simultaneous exploration of graph topology and reduction of domain disparities remains a substantial hurdle.
In this paper, we introduce the Dual Adversarial Graph Representation Learning (DAGRL), which explore the graph topology from dual branches and mitigate domain discrepancies via dual adversarial learning. Our method encompasses a dual-pronged structure, consisting of a graph convolutional network branch and a graph kernel branch, which enables us to capture graph semantics from both implicit and explicit perspectives.
Moreover, our approach incorporates adaptive perturbations into the dual branches, which align the source and target distribution to address domain discrepancies. Extensive experiments on a wild range graph classification datasets demonstrate the effectiveness of our proposed method.

\end{abstract}

\section{Introduction}
\label{sec:intro}

Graph neural networks (GNNs) have achieved remarkable accomplishments in various domains such as social network analysis~\cite{chen2020multi,gao2019geometric}, time series prediction~\cite{yin2023messages,yin2022dynamic}, and protein property prediction~\cite{yin2023omg,yin2022generic}. These networks leverage the rich relational information present in graph-structured data to make accurate predictions and learn meaningful representations. However, the success of GNNs often depends on the task-specific labels, which can be costly and slow to obtain.

To address the challenge of label acquisition, researchers have explored unsupervised domain adaptation techniques for GNNs~\cite{yin2022deal,yin2023coco,yin2023gda}. These techniques aim to transfer knowledge learned from labeled source graphs to improve the learning process for target data, which lacks labeled information. By aligning the distribution of source and target data, unsupervised domain adaptation enabling the application of GNNs in scenarios where labeled target data is scarce.

Existing work on unsupervised domain adaptation for GNNs faces significant challenges. The first is how to efficiently extract source and target topological features with limited labeled data. Simply utilizing implicit topological information is not enough for predicting unlabeled target graphs. The second challenge is the domain discrepancy. The model trained on the source domain cannot directly apply to the target domain with different data distributions. Moreover, the extracted topological information may interfere with the domain alignment. Therefore, a third challenge is how to extract graph topology exploration and reduce domain discrepancy alternately.

In this paper, we propose a novel approach called Dual Adversarial Graph Representation Learning (DAGRL) to tackle these challenges. DAGRL introduces a dual-brunch graph structure, comprising a graph convolutional network branch and a graph kernel branch. This dual-branch architecture enables us to capture graph semantics from both implicit and explicit perspectives, leveraging both the local neighborhood information and global structural properties.
Moreover, our method incorporates adaptive perturbations into the dual branches to align the source and target distributions and address domain discrepancies. By perturbing the graph representations in a controlled manner, DAGRL encourages the network to learn robust and domain-invariant features. This adversarial learning framework allows us to jointly optimize the graph topology exploration and domain adaptation iteratively, resulting in improved representation learning for target graphs.
To evaluate the effectiveness of our proposed DAGRL method, we conduct extensive experiments on a wide range of graph classification datasets. The experimental results demonstrate the superiority of our approach compared to existing state-of-the-art methods, highlighting its potential to enhance graph representation learning in various domains. The main contributions can be summarized as follows:

\begin{itemize}[leftmargin=*]
\item We propose a novel technique \method{} for graph classification that adapts to different domains. \method{} contains a graph convolutional network branch and a graph kernel network branch to capture topological information from different perspectives.
\item We introduce adaptive perturbations that align the source and target domains by perturbing the graph representations in a controlled manner, reducing domain discrepancy. Besides, we present an adversarial learning framework that alternately optimizes graph topology exploration and domain adaptation, improving the representation learning performance for target graphs.
\item Extensive experiments on various graph classification datasets demonstrate the effectiveness of the proposed \method{}.
\end{itemize}

\section{METHODOLOGY}
\label{sec:format}

\begin{figure}[t]
  \centering
\includegraphics[scale=0.66]{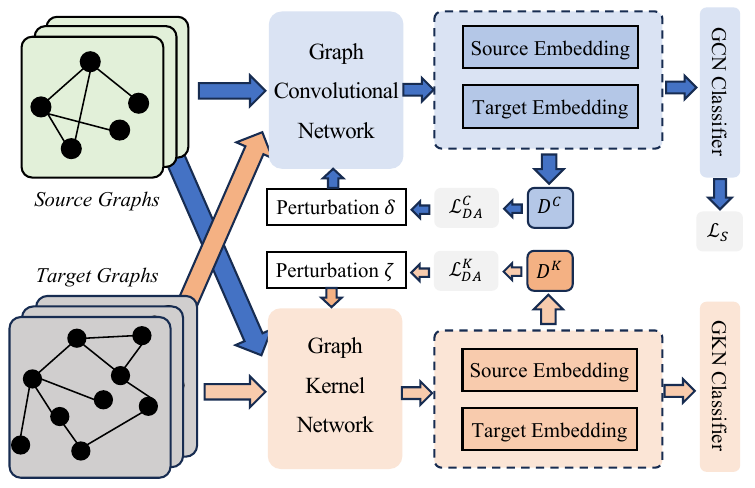}
  \caption{An overview of the proposed \method{}. The dual graph branches apply the Graph Convolutional Network and Graph Kernel Network for implicit and explicit representation. The dual adversarial perturbation learning aims at align the source and target domain shift.} 
  \label{fig1}
\end{figure}

\subsection{Problem Definition}
A graph is denoted as $G = (\mathcal{V}, \mathcal{E})$, where $\mathcal{V}$ is the set of nodes and $\mathcal{E}$ is the set of edges. The graph node feature is $\bm{X} \in \mathbb{R}^{|V| \times d}$, each row $\bm{x}_v \in \mathbb{R}^{d}$ represents the feature vector of node $v \in \mathcal{V}$ and $d$ is the dimension of node features, $|\mathcal{V}|$ is the number of nodes.
In our settings, we are given a labeled source domain $\mathcal{D}^s = \{(G_i^s, y_i^s)\}_{i=1}^{N_s}$ containing $N_s$ labeled samples and an unlabeled target domain $\mathcal{D}^t = \{G_j^t\}_{j=1}^{N_t}$ with $N_t$ examples. $\mathcal{D}^s$ and $\mathcal{D}^t$ sharing the same label space $\mathcal{Y} =\{1,2,\cdots, C\}$ but have distinct data distributions in the graph space. Our purpose is to train the graph classification model on the labeled $\mathcal{D}^s$ and unlabeled $\mathcal{D}^t$ to obtain the competitive result on the target domain.

\subsection{Dual Graph Branches}
\textbf{Graph Convolutional Network Branch}.
We use a graph convolutional network to implicitly capture the topology of the graph. For each node, we combine the embeddings of its neighbors from the previous layer. Then we update the node embedding by merging it with the combined neighbor embedding. The embedding of node $v$ at layer $l$ is $\bm{h}_v^{(l)}$:
\begin{equation}
\bm{h}_{v}^{(l)}= \operatorname{COM}^{(l)}_{\theta}\left(\bm{h}_{v}^{(l-1)}, \operatorname{AGG}^{(l)}_{\theta} \left(\left\{\bm{h}_{u}^{(l-1)}\right\}_{u \in \mathcal{N}(v)}\right) \right), \nonumber
\end{equation}
where $\mathcal{N}(v)$ denotes the neighbors of $v$. $\operatorname{AGG}^{(l)}_{\theta}$ and $\operatorname{COM}^{(l)}_{\theta}$ denote the aggregation and combination operations parameterized by $\theta$ at the $l$-th layer, respectively. 
Then, we utilize the $\operatorname{READOUT}$ function to summarize the node representations, and introduce a multilayer perception (MLP) classifier $H(\cdot)$ for final prediction:
\begin{equation}
z=F(G)=\operatorname{READOUT}\left(\bm{h}_{v}^{(L)}\right),\; \bm{p}=H(z),
\end{equation}
where $z$ denotes the graph-level representation and $\bm{p}$ is final prediction. 

\textbf{Graph Kernel Network Branch}.
Given two graph samples $G_1=(V_1,E_1)$ and $G_2=(V_2, E_2)$, graph kernels calculate their similarity by comparing their local-substructure using a kernel function. In formulation, 
\begin{equation}\label{eq:kernel}
K\left(G_{1}, G_{2}\right)=\sum_{v_{1} \in V_{1}} \sum_{v_{2} \in V_{2}} \kappa\left(l_{G_{1}}\left(v_{1}\right), l_{G_{2}}\left(v_{2}\right)\right)
\end{equation}
where $l_{G_1}(v_1)$ represents the local substructure centered at node $v_1$ and $\kappa(\cdot, \cdot)$ is a pre-defined similarity measurement. We omit $l_G(\cdot)$ and leave $\kappa(u_1,u_2)$ in Eq. \ref{eq:kernel} for simplicity.  

In this part, we apply the Weisfeiler-Lehmah (WL) Subtree Kernels as the default kernel branch. WL subtree kernels compare all subtree patterns with limited depth rooted at every node. Given the maximum depth $l$, we have:
\begin{equation}
\begin{aligned} 
 K_{\text {subtree }}^{(i)}\left(G_{s}^i, G_{t}^j\right) &=\sum_{v_{1} \in V_{s}^i} \sum_{v_{2} \in V_{t}^j} \kappa_{\text {subtree }}^{(i)}\left(u_{1}, u_{2}\right) 
\\ K_{W L}\left(G_{s}^i, G_{t}^j\right) &=\sum_{i=0}^{l} K_{\text {subtree }}^{(i)}\left(G_{s}^i, G_{t}^j\right) \end{aligned}
\end{equation}
where $\kappa_{\text {subtree }}^{(i)}\left(u_{1}, u_{2}\right)$ is derived by counting matched subtree pairs of depth $i$ rooted at node $u_{1}$ and $u_{2}$, respectively. We would assign the source label to the similar target graphs.

\begin{table*}[t]
\footnotesize
\centering
\tabcolsep=0.8pt
\caption{The classification results (in \%) on Mutagenicity under edge density domain shift (source$\rightarrow$target). M0, M1, M2, and M3 denote the sub-datasets partitioned with edge density. \textbf{Bold} results indicate the best performance.}
\resizebox{\textwidth}{!}{
\begin{tabular}{l|c|c|c|c|c|c|c|c|c|c|c|c|c}
\toprule
{\bf Methods} &M0$\rightarrow$M1 &M1$\rightarrow$M0 &M0$\rightarrow$M2 &M2$\rightarrow$M0 &M0$\rightarrow$M3 &M3$\rightarrow$M0 &M1$\rightarrow$M2 &M2$\rightarrow$M1 &M1$\rightarrow$M3 &M3$\rightarrow$M1 &M2$\rightarrow$M3 &M3$\rightarrow$M2 &Avg.\\
\midrule
WL subtree  &74.9 &74.8 &67.3 &69.9 &57.8 &57.9 &73.7 &80.2 &60.0 &57.9 &70.2 &73.1 &68.1\\
GCN &71.1 &70.4 &62.7 &69.0 &57.7 &59.6 &68.8 &74.2 &53.6 &63.3 &65.8 &74.5 &65.9\\
GIN &72.3 &68.5 &64.1 &72.1 &56.6 &61.1 &67.4 &74.4 &55.9 &67.3 &62.8 &73.0 &66.3\\
CIN &66.8 &69.4 &66.8 &60.5 &53.5 &54.2 &57.8 &69.8 &55.3 &74.0 &58.9 &59.5 &62.2\\
GMT &73.6 &75.8 &65.6 &73.0 &56.7 &54.4 &72.8 &77.8 &62.0 &50.6 &64.0 &63.3 &65.8\\
\midrule
CDAN &73.8 &74.1 &68.9 &71.4 &57.9 &59.6 &70.0 &74.1 &60.4 &67.1 &59.2 &63.6 &66.7\\
ToAlign &74.0 &72.7 &69.1 &65.2 &54.7 &73.1 &71.7 &77.2 &58.7 &73.1 &61.5 &62.2 &67.8\\
MetaAlign &66.7 &51.4 &57.0 &51.4 &46.4 &51.4 &57.0 &66.7 &46.4 &66.7 &46.4 &57.0 &55.4\\
\midrule
DEAL &76.3 &72.6 &69.8 &73.3  &58.3 &71.2 &\textbf{77.9} &80.8 &64.1 &74.1 &70.6 &74.9 &72.0\\
CoCo &77.7 &76.6  &\textbf{73.3} &74.5 &66.6 &74.3 &77.3 &80.8 &67.4 &74.1 &68.9 &77.5 &74.1\\
\midrule 
\method{}   &\textbf{77.9} &\textbf{76.8} &72.8 &\textbf{75.3} &\textbf{66.7} &\textbf{75.4} &77.8 &\textbf{81.0} &\textbf{67.7} &\textbf{75.3} &\textbf{71.1} &\textbf{77.7} &\textbf{74.6}\\
\bottomrule
\end{tabular}}
\label{tab::results}
\end{table*}

\begin{table*}[t]
\footnotesize
\centering
\tabcolsep=3pt
\caption{The classification results (in \%) on Tox21 under edge density domain shift (source$\rightarrow$target). T0, T1, T2, and T3 denote the sub-datasets partitioned with edge density. \textbf{Bold} results indicate the best performance.}
    \resizebox{\textwidth}{!}{
\begin{tabular}{l|c|c|c|c|c|c|c|c|c|c|c|c|c}
\toprule
{\bf Methods} &T0$\rightarrow$T1 &T1$\rightarrow$T0 &T0$\rightarrow$T2 &T2$\rightarrow$T0 &T0$\rightarrow$T3 &T3$\rightarrow$T0 &T1$\rightarrow$T2 &T2$\rightarrow$T1 &T1$\rightarrow$T3 &T3$\rightarrow$T1 &T2$\rightarrow$T3 &T3$\rightarrow$T2 &Avg.\\
\midrule
WL subtree  &65.3 &51.1 &69.6 &52.8 &53.1 &54.4 &71.8 &65.4 &60.3 &61.9 &57.4 &76.3 &61.6\\
GCN &64.2 &50.3 &67.9 &50.4 &52.2 &53.8 &68.7 &61.9 &59.2 &51.4 &54.9 &76.3 &59.3\\
GIN &67.8 &51.0 &77.5 &54.3 &56.8 &54.5 &78.3 &63.7 &56.8 &53.3 &56.8 &77.1 &62.3\\
CIN &67.8 &50.3 &78.3 &54.5 &56.8 &54.5 &78.3 &67.8 &59.0 &67.8 &56.8 &78.3 &64.2\\
GMT &67.8 &50.0 &78.4 &50.1 &56.8 &50.7 &78.3 &67.8 &56.8 &67.8 &56.4 &78.1 &63.3\\
\midrule
CDAN &69.9 &55.2 &78.3 &56.0 &59.5 &56.6 &78.3 &\textbf{68.5} &61.7 &68.1 &61.0 &78.3 &66.0\\
ToAlign &68.2 &58.5 &78.4 &58.8 &58.5 &53.8 &\textbf{78.8} &67.1 &64.4 &68.8 &57.9 &78.4 &66.0\\
MetaAlign &65.7 &57.5 &78.0 &58.5 &\textbf{63.9} &52.2 &78.8 &67.1 &62.3 &67.5 &56.8 &78.4 &65.6\\
\midrule
DEAL &73.9 &59.4 &68.2 &58.2 &53.8 &58.7 &69.2 &66.2 &63.5 &67.4 &61.1 &77.8 &64.8\\
CoCo &69.9 &59.8 &78.8  &59.0 &62.3 &59.0 &78.4 &66.8 &65.0 &68.8 &61.2 &78.4 &67.3\\
\midrule 
\method{}   &\textbf{74.2} &\textbf{60.1} &\textbf{78.8} &\textbf{59.3} &62.7 &\textbf{59.4} &78.1 &67.2 &\textbf{65.7} &\textbf{69.3}    &\textbf{62.8} &\textbf{79.2} &\textbf{68.1}\\
\bottomrule
\end{tabular}}
\label{tab::results1}
\end{table*}

\subsection{Dual Adversarial Perturbation Learning}
In this part, our goal is to use perturbations on source graphs to mitigate the impact of domain inconsistencies on feature representations with target semantics. We employ adversarial training to identify promising perturbation directions. Taking the GCN branch as an example, we obtain feature representations and label predictions for both the source and target domains. We train a domain discriminator $D(\cdot)$ on each branch to differentiate between the two domains. For each source graph $G^s_i$, we introduce noise $\bm{\delta}(\bm{H}^s_i)$ to the node embedding $\bm{H}^s_i$. The direction of perturbation is determined through a minimax optimization process, and $\bm{\delta}$ is updated based on the gradient descent direction for each source graph $G^s_i$.
\begin{align}
\label{eq:per_1}
 \min_{||\delta(\cdot)||_F\leq \epsilon} \max_{\theta_{d}} 
\mathcal{L}_{DA}^{C}&=  \mathbb{E}_{G^s_i \in \mathcal{D}^s} \log D(F(G^s_i; \bm{H}^s_i + \bm{\delta}(\bm{H}^s_i)), {\bm{p}}^s_i)  \nonumber\\ &+   \mathbb{E}_{G^t_j\in \mathcal{D}^t} \log (1-D(F(G^t_j), {\bm{p}}^t_j)), \nonumber \\
\boldsymbol{\delta}_{t+1}&=\boldsymbol{\delta}_{t}-\epsilon
\phi\left(\boldsymbol{\delta}_{t}\right) /\left\|\phi\left(\boldsymbol{\delta}_{t}\right)\right\|_{F},
\end{align}
where $\theta_{d}$ is the parameters of the domain discriminator and $\epsilon$ is the maximum of perturbation size. $\phi\left(\boldsymbol{\delta}_{t}\right)=\nabla_{\boldsymbol{\delta}} \log D(F(G^s_i; \bm{H}^s_i + \bm{\delta}(\bm{H}^s_i)))$ is the gradient of the loss for $G^s_i$ with respect to the perturbation $\bm{\delta}$. Similarity, we can get the perturbation $\boldsymbol{\zeta}$ on the GKN branch by optimize the minimax objective, i.e., $\min_{||\zeta(\cdot)||_F\leq \epsilon} \max_{\theta_{d}'} 
\mathcal{L}_{DA}^{K}$.

\subsection{Learning Framework}
The primary training objective of \method{} combines adversarial perturbation loss and target data classification loss. Additionally, minimizing the expected source error for labeled source samples is essential. 
\begin{equation}
\mathcal{L}_{S}=\mathbb{E}_{G_i^s \in \mathcal{D}^s} \mathcal{E} (H(\hat{\bm{z}}_i^s), y_i^s),
\end{equation}
where $\hat{\bm{z}}_i^s$ denotes the representations of source data under perturbations.
Consequently, to update the network parameters, we minimize the ultimate objective as follows:
\begin{equation}\label{eq:final}
    \mathcal{L} = \mathcal{L}_{S} - \lambda_1 \mathcal{L}_{DA}^C-\lambda_2 \mathcal{L}_{DA}^K,
\end{equation}
where $\lambda_1$ and $\lambda_2$ are hyper-parameters to balance the domain adversarial loss and classification loss. 
Meanwhile, we need to update the perturbations of source samples, i.e., $\boldsymbol{\delta}$ and $\boldsymbol{\zeta}$, and then update the model parameters interactively.


\begin{table*}[t]
\footnotesize
\centering
\tabcolsep=0.5pt
\caption{The results of ablation studies on Mutagenicity (source$\rightarrow$target). }\label{tab::ablation}
\resizebox{\textwidth}{!}{
\begin{tabular}{l|c|c|c|c|c|c|c|c|c|c|c|c|c}
\toprule
{\bf Methods} &M0$\rightarrow$M1 &M1$\rightarrow$M0 &M0$\rightarrow$M2 &M2$\rightarrow$M0 &M0$\rightarrow$M3 &M3$\rightarrow$M0 &M1$\rightarrow$M2 &M2$\rightarrow$M1 &M1$\rightarrow$M3 &M3$\rightarrow$M1 &M2$\rightarrow$M3 &M3$\rightarrow$M2 &Avg.\\
\midrule
\method{}/P1 &76.3 &72.6 &64.7 &73.2 &64.1 &73.3 &76.4 &78.6 &63.7 &73.4 &68.8 &74.6 &71.6\\
\method{}/P2 &76.7 &73.9 &65.4 &73.8 &65.2 &73.4 &75.9 &78.1 &64.5 &74.4 &67.2 &75.9 &72.0\\
\method{}-GIN &77.0 &74.2 &64.6 &73.3 &64.3 &74.1 &76.3 &79.5 &66.8 &75.2 &68.7 &76.0 &72.5\\
\method{}-GKN &76.8 &73.7 &65.3 &74.1 &65.2 &74.7 &76.7 &78.6 &66.4 &74.3 &67.7 &77.5 &72.6\\
\midrule
\method{}   &\textbf{77.9} &\textbf{76.8} &\textbf{72.8} &\textbf{75.3} &\textbf{66.7} &\textbf{75.4} &\textbf{77.8} &\textbf{81.0} &\textbf{67.7} &\textbf{75.3} &\textbf{71.1} &\textbf{77.7} &\textbf{74.6}\\

\bottomrule 
\end{tabular}
}
\vspace{-0.3cm}
\end{table*}

\section{Experiments}

\subsection{Datasets and Baselines}
\textbf{Datasets}.
To evaluate the proposed method's effectiveness, we conducted experiments using our method on several real-world datasets: Mutagenicity (M)~\cite{kazius2005derivation}, and Tox21
\footnote{https://tripod.nih.gov/tox21/challenge/data.jsp.} (T) from the TUDataset~\cite{Morris2020}. 
Mutagenicity dataset consists of 4337 molecular structures and their corresponding Ames test data. Tox21 dataset is used to assess the predictive ability of models in detecting compound interferences.
To address domain distribution variations within each dataset, we divided them into four sub-datasets: $D$0, $D$1, $D$2, and $D$3 ($D$ represents each respective dataset) based on edge density~\cite{yin2023coco}.

\textbf{Baselines}. 
To evaluate the effectiveness of our method, we compare the proposed method with a large number of state-of-the-art methods, including five graph approaches ({WL subtree}~\cite{shervashidze2011weisfeiler}, {GCN}~\cite{kipf2017semi}, {GIN}~\cite{xu2019powerful}, {CIN}~\cite{bodnar2021weisfeiler} and {GMT}~\cite{BaekKH21}), and three recent domain adaptation methods ({CDAN}~\cite{long2018conditional}, {ToAlign}~\cite{NeurIPS2021_731c83db}, {MetaAlign}~\cite{wei2021metaalign}) and two graph domain adaptation methods (DEAL~\cite{yin2022deal}, CoCo~\cite{yin2023coco}).

\subsection{Implementation Details}
In our implementation, a two-layer GIN~\cite{xu2019powerful} is employed in the GCN branch, while a two-layer network with the Weisfeiler-Lehman (WL) kernel \cite{shervashidze2011weisfeiler} is used in the GKN branch. The Adam optimizer with a learning rate of $10^{-4}$ is used by default. Both the embedding dimension of hidden layers and the batch size are set to 64.

\subsection{Performance Comparison}
Table~\ref{tab::results} and \ref{tab::results1} display the performance of graph classification in different unsupervised domain adaptation scenarios. The results indicate the following observations:
1) Domain adaptation methods outperform graph kernel and GNN methods, suggesting that current graph classification models lack transfer learning capabilities. Therefore, an effective domain adaptive framework is crucial for graph classification.
2) While domain adaptation methods achieve competitive results in simple transfer tasks, they struggle to make significant progress in challenging transfer tasks compared to GIN. This difficulty is attributed to the complexity of acquiring graph representations, making direct application of current domain adaptation approaches to GCNs unwise.
3) \method{} achieves superior performance compared to all baselines in most cases. This improvement is attributed to two key factors: (i) Dual branches (GCN and GKN) enhances graph representation in situations with limited labels. (ii) Adversarial perturbation learning facilitates effective domain alignment.

\subsection{Ablation Study}
To evaluate the impact of each component in our \method{}, we introduce several model variants as follows: 
1) \textbf{\method{}}/\textbf{P1}: It removes the perturbation on the GCN branch;
2) \textbf{\method{}}/\textbf{P2}: It removes the perturbation on the GKN branch; 
3) \textbf{\method{}-GIN}: It uses two different GINs to generate graph representations; 
4) \textbf{\method{}-GKN}: It uses two different GKN to generate graph representations.

We conducted performance comparisons on Mutagenicity, and the results are presented in Table~\ref{tab::ablation}. Based on the observations from the table, the following findings can be summarized:
1) \method{}/P1 and \method{}/P2 exhibit worse performance compared to \method{}. This confirms that adaptive adversarial perturbation learning is effective in achieving domain alignment.
2) Both \method{}-GIN and \method{}-GKN produce similar results but perform worse than \method{}. This suggests that relying solely on ensemble learning in either implicit or explicit manner does not significantly enhance graph representation learning when labels are scarce.

\section{Conclusion}
We introduce the practical problem of unsupervised domain adaptive graph classification named \method{}. \method{} is proposed with two branches, i.e., a graph convolutional network branch and a graph kernel network branch, which explores graph topological information in implicit and explicit manners, respectively. Then, we introduce the dual adaptive adversarial perturbation learning to minimize the domain discrepancy. Throughout the training process, we iteratively update the perturbation direction and model parameters, enabling us to align domain distributions and learn graph representations more accurately. Extensive experiments valdate the effectiveness of the proposed \method{}.


\bibliographystyle{IEEEtran}
\bibliography{main}

\end{document}